\documentclass[10pt,twocolumn,letterpaper]{article}

\usepackage{iccv}              

\usepackage{url}
\usepackage{booktabs}
\usepackage{multirow}
\usepackage{graphicx}
\usepackage{xspace}

\usepackage{mathtools}
\usepackage{adjustbox}
\usepackage{array}
\usepackage{colortbl}
\usepackage{float,wrapfig}
\usepackage{hhline}
\usepackage{subcaption} 
\usepackage{enumitem}

\usepackage{tikz}
\usetikzlibrary{tikzmark}
\usepackage{pifont}

\usepackage{wrapfig}

\newcolumntype{g}{>{\columncolor{mitblue}}c}

\newcolumntype{i}{>{\columncolor{gray}}c}

\definecolor{cvprblue}{rgb}{0.21,0.49,0.74}
\definecolor{mitblue}{rgb}{0.88,0.95,0.96}
\definecolor{gold}{rgb}{0.75,0.6,0.12}
\colorlet{shadecolor}{gray!40}
\definecolor{mydarkred}{rgb}{0.8,0.02,0.02}

\usepackage{amsmath,amsfonts,bm}

\newcommand{\cmark}{\ding{51}}%
\newcommand{\xmark}{\ding{55}}%

\def\modelfull{DC-AE 1.5\xspace}
\def\modelshort{DC-AE 1.5\xspace}
\def\modelterm{DC-AE-1.5\xspace}

\def\aetech{Structured Latent Space\xspace}
\def\diffusiontech{Augmented Diffusion Training\xspace}

\makeatletter
\def\blfootnote#1{\xdef\@thefnmark{}\@footnotetext{\scriptsize #1}}
\makeatother

\definecolor{iccvblue}{rgb}{0.21,0.49,0.74}
\usepackage[pagebackref,breaklinks,colorlinks,allcolors=iccvblue]{hyperref}

\def\paperID{14802} 
\def\confName{ICCV}
\def\confYear{2025}

\title{
\modelshort: Accelerating Diffusion Model Convergence \\ with Structured Latent Space
}

\author{
Junyu Chen \quad 
Dongyun Zou \quad
Wenkun He \quad 
Junsong Chen \quad 
Enze Xie \quad 
Song Han \quad 
Han Cai \\
NVIDIA \\
\url{https://github.com/dc-ai-projects/DC-Gen}
}

\begin{document}
\maketitle

\blfootnote{Correspondence to: Han Cai (\texttt{hcai@nvidia.com}).}

\begin{abstract}
We present \modelfull, a new family of deep compression autoencoders for high-resolution diffusion models. Increasing the autoencoder's latent channel number is a highly effective approach for improving its reconstruction quality. However, it results in slow convergence for diffusion models, leading to poorer generation quality despite better reconstruction quality. This issue limits the quality upper bound of latent diffusion models and hinders the employment of autoencoders with higher spatial compression ratios. 
We introduce two key innovations to address this challenge: i) \textbf{\aetech}, a training-based approach to impose a desired channel-wise structure on the latent space with front latent channels capturing object structures and latter latent channels capturing image details; ii) \textbf{\diffusiontech}, an augmented diffusion training strategy with additional diffusion training objectives on object latent channels to accelerate convergence. 
With these techniques, \modelfull delivers faster convergence and better diffusion scaling results than DC-AE. On ImageNet 512$\times$512, \modelterm-f64c128 delivers better image generation quality than DC-AE-f32c32 while being 4$\times$ faster. 
\end{abstract}
\section{Introduction}
\label{sec:intro}

\begin{figure*}[t]
    \centering
    \includegraphics[width=\linewidth]{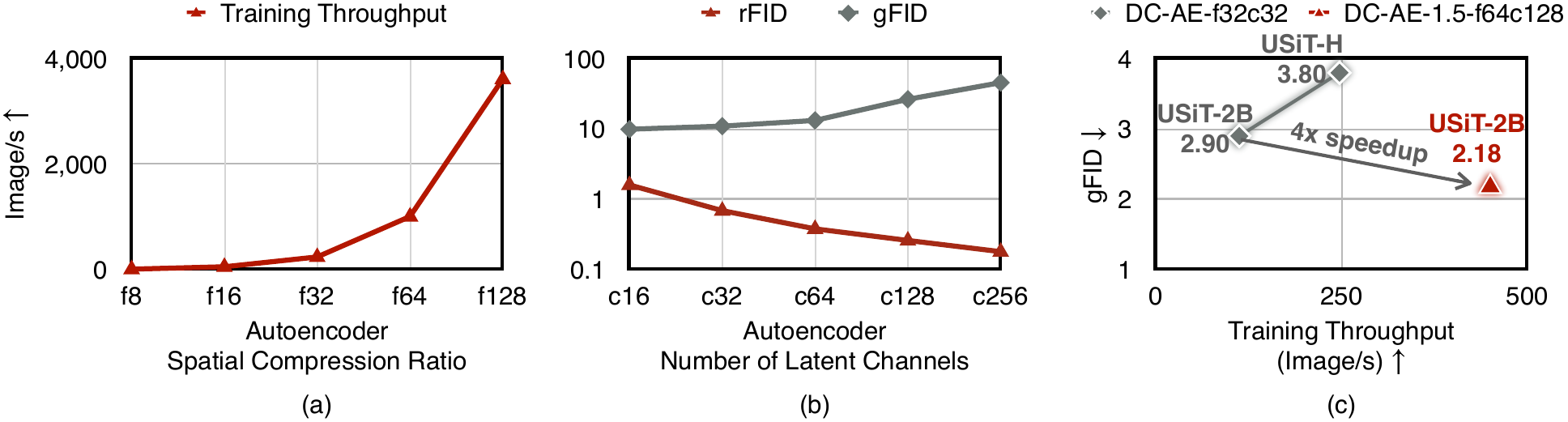}
    \caption{\textbf{(a) Training Throughput Comparison under Different Autoencoder Spatial Compression Ratios.} Increasing the autoencoder's spatial compression ratio effectively improves diffusion models' training efficiency by producing a latent space with fewer tokens. However, larger latent channel numbers are required to maintain satisfactory reconstruction quality. \textbf{(b) rFID and gFID Results under Different Latent Channel Numbers.} We use DiT-XL \cite{peebles2023scalable} as the diffusion model. rFID keeps improving with more latent channels, while gFID keeps getting worse. \textbf{(c) Efficiency-Quality Trade-off Comparison on ImageNet 512$\times$512.} Classifer-free guidance \cite{ho2022classifier} is not used. \modelterm-f64c128 delivers 4$\times$ speedup over DC-AE-f32c32 while maintaining a better image generation quality.}
    \vspace{-15pt}
    \label{fig:figure1}
\end{figure*}

Latent Diffusion Model (LDM) \cite{rombach2022high} has established itself as a dominant paradigm in high-resolution image synthesis \cite{flux2024,esser2024scaling,xie2025sana,liu2024playground}. It leverages an autoencoder to project images of size \(H\times W\times 3\) into compressed latent representations of shape \(\frac H f\times\frac W f\times c\)\footnote{`f' denotes the autoencoder's spatial compression ratio. `c' represents the latent channel number of the autoencoder.}, reducing the computational cost of the diffusion model. The diffusion model's outputs are fed to the autoencoder at inference time to reconstruct images from the latent representations. In LDM, the autoencoder reconstruction quality is critical, as it sets a quality upper bound for the whole image synthesis pipeline. 

A common approach to improve the quality of autoencoder reconstruction is to increase its latent channel number \cite{dai2023emu,esser2024scaling,flux2024}. For example, Figure~\ref{fig:figure1} (b) shows the autoencoder's rFID (reconstruction FID, the lower the better) results with different latent channel numbers. We can see that the rFID consistently improves as the latent channel number increases, decreasing from 1.60 to 0.26 if switching from c16 to c128. 

Expanding the latent channel number is especially critical for deep compression autoencoders \cite{chen2024deep}, which accelerate latent diffusion models by increasing the autoencoder's spatial compression ratio (Figure~\ref{fig:figure1}a). Under a high spatial compression ratio (e.g., f64), deep compression autoencoders must use a large latent channel number (e.g., c128) to maintain a satisfactory reconstruction quality. 

However, using a large latent channel number significantly slows the diffusion model's convergence, leading to worse gFID (generation FID, the lower the better) results. For example, Figure~\ref{fig:figure1} (b) demonstrates DiT-XL's \cite{peebles2023scalable} gFID results with different latent channel numbers. Although the autoencoder's rFID keeps improving, the gFID keeps deteriorating. This issue not only limits LDM's quality upper bound but also bounds its efficiency as it hinders the employment of autoencoders with high spatial compression ratios (e.g., f64). 

\begin{figure*}[t]
    \centering
    \includegraphics[width=0.95\linewidth]{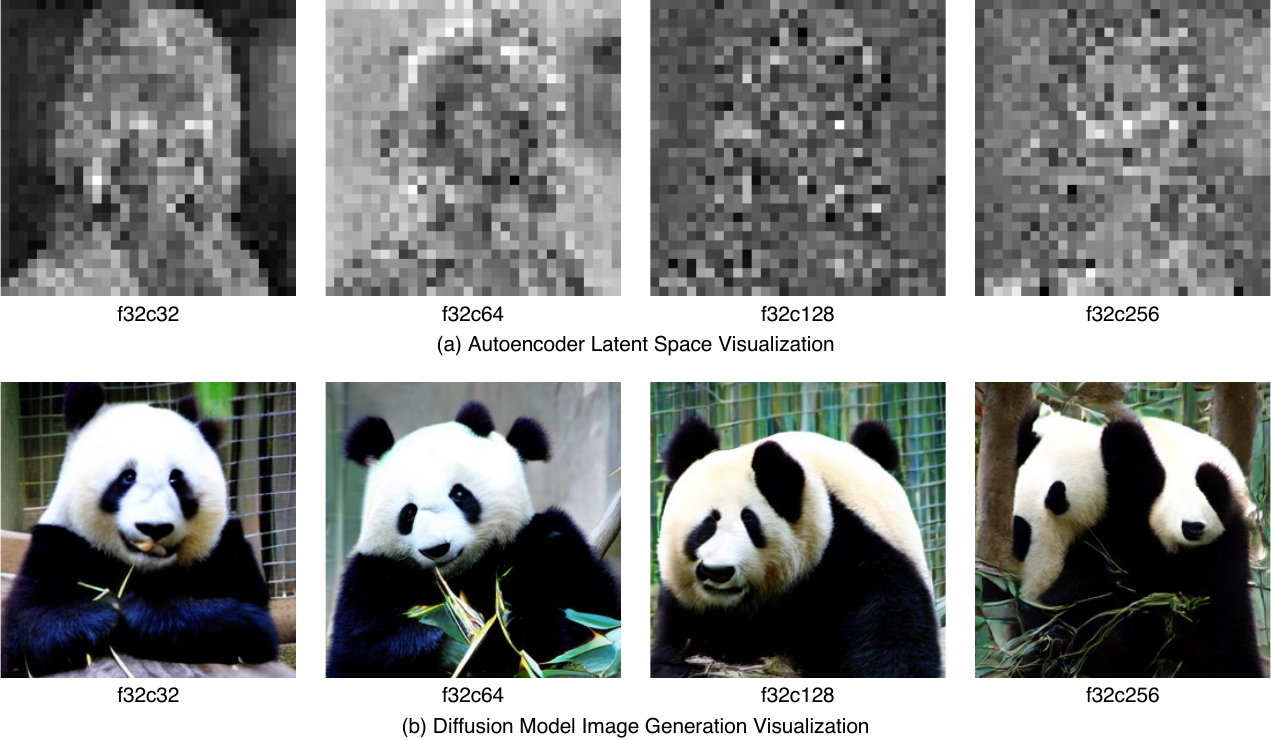}
    \caption{
    We visualize the channel-wise average feature here. We provide the complete latent space visualization in the supplementary material (Figure~\ref{fig:latent_space_visualization_1} and \ref{fig:latent_space_visualization_2}). The visualization shows that the object structure information gets blurred if we increase the latent channel number. It makes diffusion models unable to learn object structure efficiently. As a result, we can see gradually distorted object structures when we enlarge the latent channel number, as shown in the visualization of the diffusion model's outputs. We use the DiT-XL as the diffusion model here.
    }
    \vspace{-5pt}
    \label{fig:analysis_and_motivation}
\end{figure*}

This work presents \modelfull, which introduces two key innovations to tackle the aforementioned challenge. First, we analyze the autoencoder's latent space under different latent channel numbers (Section~\ref{subsec:analysis_and_motivation}). We find that the latent space has a sparsity issue when using a large number of latent channels (Figure~\ref{fig:analysis_and_motivation}a). It allocates most of the latent channels for capturing image details while the latent channels for capturing object structure become sparse among the whole latent space\footnote{We refer to them as \emph{detail latent channels} and \emph{object latent channels}, respectively.}. This sparsity issue makes it more difficult for diffusion models to learn object structure, leading to the slow convergence issue. As shown in Figure~\ref{fig:analysis_and_motivation} (b), the image details remain good when we increase the latent channel number. However, the object structures suffer from significant distortion. 

Motivated by the findings, we propose \textbf{\aetech} (Section~\ref{subsec:structured_latent_space}) to alleviate the sparsity issue. It introduces a training-based approach (Figure~\ref{fig:ae_pipeline}) to imposing a specific structure on the latent space: the front latent channels focus on capturing object structure, while the latter latent channels focus on capturing image details (Figure~\ref{fig:structured_latent_space}). Second, based on the \aetech, we propose \textbf{\diffusiontech} (Section~\ref{subsec:augmented_diffusion_training}) to address the slow convergence issue. It introduces extra diffusion training objectives on object latent channels to accelerate diffusion models' learning speed on capturing object structure (Figure~\ref{fig:augmented_diffusion_training}). 

With these techniques, we significantly accelerate diffusion models' convergence when using a large latent channel number (Figure~\ref{fig:augmented_diffusion_training}b), leading to better gFID results (Table~\ref{tab:better_convergence_imagenet_256} and~\ref{tab:better_convergence_imagenet_512}). In addition, it also leads to better diffusion model scaling results than previous autoencoders (e.g., DC-AE-f32c32), as shown in Figure~\ref{fig:better_scaling_curve}. On ImageNet 512$\times$512, \modelterm-f64c128+USiT-2B achieves 2.18 gFID without classifier-free guidance, surpassing DC-AE-f32c32+USiT-2B while being 4$\times$ faster (Figure~\ref{fig:figure1}c).
We summarize our contributions as follows:

\begin{itemize}[leftmargin=*]
\item We analyze the latent space of autoencoders under different latent channel numbers. Our study reveals the latent space's sparsity issue when using a large latent channel number, which leads to the slow convergence issue for diffusion models. 
\item We propose \aetech and \diffusiontech, which effectively address the slow convergence issue. 
\item We build \modelfull based on our key innovations. It delivers better diffusion scaling results than prior autoencoders, boosting the quality upper bound of LDMs and paving the way for the employment of autoencoders with higher spatial compression ratios. 
\end{itemize}

\section{Related Work}
\label{sec:related}

\paragraph{Autoencoders in Image Generation.}
Training diffusion models in high-resolution pixel space is computationally prohibitive due to the large token number. To overcome this challenge, \cite{rombach2022high} proposes to employ pretrained autoencoders to produce a compressed latent space and operate diffusion models on the compressed space. This strategy significantly reduces diffusion models' training and inference costs, making it possible to train large-scale diffusion models on web-scale datasets, leading to impressive image generation results \cite{flux2024,esser2024scaling,liu2024playground}. 

Following that, one line of works focuses on improving the autoencoders' reconstruction quality by increasing the latent channel number \cite{esser2024scaling,dai2023emu,flux2024} or adding task-specific priors \cite{zhu2023designing}. Another line of works focuses on improving diffusion models' efficiency by increasing autoencoders' spatial compression ratio \cite{chen2024deep,chen2024softvq,chen2025masked}. A third line of work \cite{vahdat2020nvae,child2020very,shu2022bit,takida2023hq} explores hierarchical VAEs with multi-resolution latent spaces, while our approach focuses on introducing structure in the latent channel dimension. Concurrent to our work, some recent works propose to make the latent space more friendly for diffusion models by aligning it with a pretrained discriminative model \cite{yao2025reconstruction} or leveraging scale equivariance regularization \cite{skorokhodov2025improving,kouzelis2025eq}. 
Unlike prior works, our work addresses the slow convergence issue of diffusion models when using a large latent channel number (e.g., c128). It also provides novel insights and solutions to this critical problem. 

\paragraph{Accelerating Diffusion Model Convergence.}
Training the diffusion model consumes extensive computation resources, motivating many works to improve its training efficiency. Representative techniques include designing better model architectures \cite{bao2023all,cai2024condition,cai2023efficientvit}, representation alignment with pretrained discriminative models \cite{yu2024representation}, better diffusion training schedulers \cite{ma2024sit,esser2024scaling}, improving the training data quality \cite{chenpixart,xie2024sana}, etc. Orthogonal to these techniques, our work accelerates diffusion model convergence by employing a structured latent space. 

\section{Method}
\label{sec:method}

In Section~\ref{subsec:analysis_and_motivation}, we first analyze the latent spaces of autoencoders with different latent channel numbers to understand the underlying source of the slow convergence issue. Next, in Section~\ref{subsec:structured_latent_space} and Section~\ref{subsec:augmented_diffusion_training}, we introduce \modelfull with \emph{\aetech} and \emph{\diffusiontech} to address the slow convergence issue. 

\subsection{Analysis and Motivation}
\label{subsec:analysis_and_motivation}

\begin{figure*}[htbp]
    \centering
    \includegraphics[width=\textwidth]{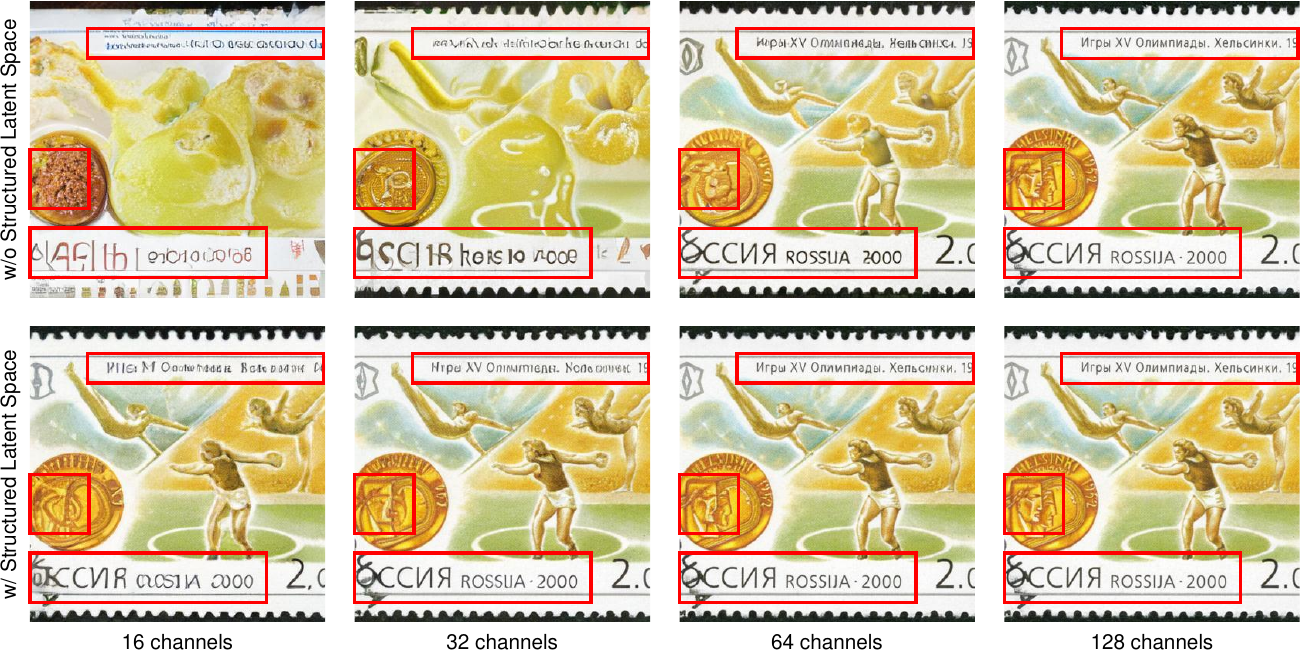}
    \caption{\textbf{Image Reconstruction Comparison.} With the structured latent space, \modelfull can reconstruct images given partial latent channels, with front latent channels reconstructing overall object structure and semantics and latter latent channels adding details. In contrast, DC-AE can not reconstruct the object structure well given partial latent channels. The decoder is fine-tuned for all settings to fully reveal the information encoded in the (partial) latent space. } 
    \vspace{-5pt}
    \label{fig:structured_latent_space}
\end{figure*}
\begin{figure}[t]
    \centering
    \includegraphics[width=\linewidth]{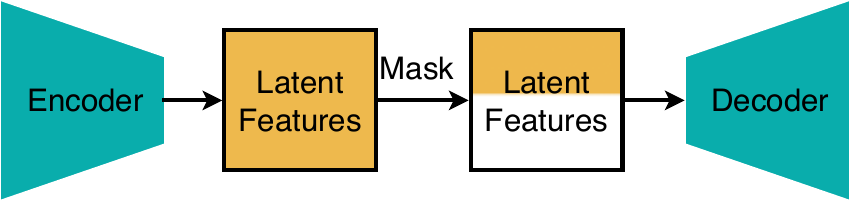}
    \caption{\textbf{Illustration of \modelfull Autoencoder Training Strategy.} The key difference from conventional autoencoder training is that we add a channel-wise random masking step before feeding the latent features to the decoder. The mask is generated randomly at each step according to Eq.~\ref{eq:mask}. It enables the autoencoder to reconstruct with partial latent channels and naturally impose the channel-wise structure on the latent space. } 
    \vspace{-5pt}
    \label{fig:ae_pipeline}
\end{figure}

We visualize the latent spaces of DC-AE-f32 \cite{chen2024deep} with different latent channels (c32, c64, c128, and c256) to analyze why diffusion models suffer from the slow convergence issue when using a large latent channel number. Due to the page limit, we provide the complete latent space visualization in the supplementary material. In Figure~\ref{fig:analysis_and_motivation} (a), we provide the visualization of the channel-wise average feature of the latent spaces\footnote{Assuming each latent representation has a shape of $[h, w, c]$, we get its channel-wise average feature via `torch.mean(latent, dim=2)'.}. 

We can see that the information about the object structure becomes more sparse as we increase the latent channel number from c32 to c256. From the complete latent space visualization (Figure~\ref{fig:latent_space_visualization_1}), we find that this phenomenon is because the f32c256 autoencoder contains a lot of latent channels for capturing image details. In contrast, only a few latent channels are responsible for capturing object structure. 

It is reasonable since capturing more image details is critical for achieving a good reconstruction quality. However, since we treat all latent channels equally, it is challenging for diffusion models to distinguish between the overall object structure and high-frequency details, making them unable to learn object structure efficiently. For example, we visualize the image generation results of DiT-XL \cite{peebles2023scalable} with different autoencoders (DC-AE-f32c32 $\rightarrow$ DC-AE-f32c256) in Figure~\ref{fig:analysis_and_motivation} (b). We can see that the diffusion model gradually loses control over structural coherence as the number of latent channels increases. In contrast, the image details remain good under all settings. 

Based on these findings, we conjecture \emph{the autoencoder's latent space suffers from the object information sparsity issue as demonstrated in Figure~\ref{fig:analysis_and_motivation} (a) when using a large latent channel number. This sparsity issue makes diffusion models unable to learn object structure efficiently, leading to slow convergence.}

\begin{figure*}[t]
    \centering
    \includegraphics[width=\linewidth]{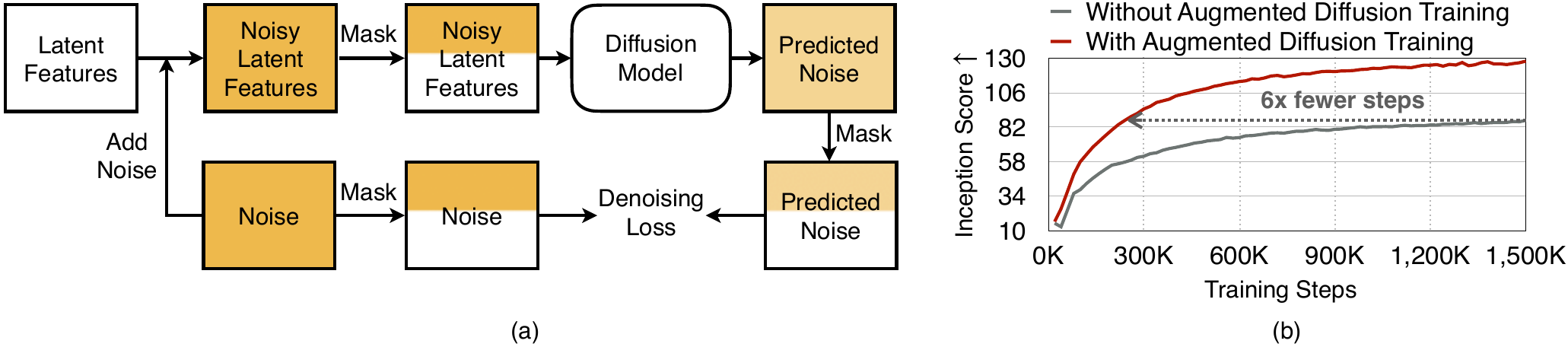}
    \vspace{-5pt}
    \caption{\textbf{(a) Illustration of \diffusiontech.} We randomly generate a channel-wise mask at each training step and use it to augment diffusion training. \textbf{(b) Training Curve Comparison.} We achieve 6$\times$ faster convergence on UViT-H \cite{bao2023all} with \diffusiontech.}
    \vspace{-5pt}
    \label{fig:augmented_diffusion_training}
\end{figure*}

\subsection{Structured Latent Space}
\label{subsec:structured_latent_space}

Motivated by the above analysis, we introduce \aetech to help diffusion models distinguish object latent channels from detail latent channels to tackle the latent space sparsity issue. Figure~\ref{fig:structured_latent_space} illustrates the general idea. Unlike the conventional autoencoders, whose latent spaces have no structure along the latent channel dimension, we design and add a channel-wise structure on the latent space. 

Specifically, \modelfull has an additional capacity of reconstructing images from partial latent channels (e.g., first 16/32/64 channels from c128). For example, as shown in Figure~\ref{fig:structured_latent_space}, it first focuses on reconstructing object structures from the first 16 latent channels and gradually refines image details as we include more latent channels.
In contrast, when given partial latent channels, the conventional autoencoder (e.g., DC-AE \cite{chen2024deep}) cannot reconstruct the object structure well. 

In the supplementary material, we visualize the complete latent space comparison between DC-AE and \modelfull (Figure~\ref{fig:latent_space_visualization_2}). We can see a clear separation between object latent channels and detail latent channels in \modelfull's latent space with front channels as object latent channels and latter channels as detail latent channels, which is absent in DC-AE's latent space. 

Figure~\ref{fig:ae_pipeline} demonstrates our autoencoder training strategy to achieve the desired channel-wise latent space structure. The intuition of our design is based on the finding that the autoencoder's latent space naturally focuses more on the object structure when the latent channel number is small. Therefore, we augment the original autoencoder training objective with the additional objective of reconstructing input images from partial latent channels.  

Specifically, an autoencoder consists of an encoder \(E\) that maps the input image \(x\in \mathbb R^{H\times W\times 3}\) to the latent feature \(z = E(x)\in \mathbb R^{\frac H f\times \frac W f\times c}\) and a decoder \(D\) that predicts the image \(y= D(z)\in\mathbb R^{H\times W\times 3}\) from the latent feature. A training loss \(l(x, y)=l(x, D(z))\) is used to supervise the autoencoder's training, which is a weighted average of the L1 loss, the perceptual loss \cite{zhang2018unreasonable}, and the GAN loss \cite{isola2017image}.

In our design, we also require the autoencoder to reconstruct the image when only the first several channels are selected. In practice, we implement this by randomly sampling a latent channel number $(c'\le c, c' \in [c_1, c_2, \cdots, c])$ at each training step and generate a mask
\begin{equation}
    \mathrm{mask}_{c,c'}=(\underbrace{1, 1, \dots, 1}_{c'}, \underbrace{0, 0, \dots, 0}_{c-c'}).\label{eq:mask}
\end{equation}
Then we use the modified loss \(l(x, D(z\cdot \mathrm{mask}_{c,c'}))\) to update the autoencoder.

Through this training strategy, the autoencoder gains the ability to reconstruct images given any latent channel number $c' \in [c_1, c_2, \cdots, c]$, and its latent space naturally has our desired channel-wise structure. In addition, this training strategy has little impact on the autoencoder's reconstruction quality. Table~\ref{tab:rfid_results} shows the rFID comparison between DC-AE and \modelfull. \modelfull achieves the same rFID as DC-AE under the same setting while having the additional latent space structure. 

\begin{table}[t]
\small\centering\setlength{\tabcolsep}{4pt}
\begin{tabular}{l | c c | c }
\toprule
\multirow{2}{*}{Autoencoder} & \multirow{2}{*}{Latent Shape} & Structured & \multirow{2}{*}{rFID $\downarrow$} \\
&  & Latent Space & \\
\midrule
DC-AE-f32c32 \cite{chen2024deep} & 8$\times$8$\times$32 & \xmark & 0.69 \\
DC-AE-f32c128 \cite{chen2024deep} & 8$\times$8$\times$128 & \xmark & 0.26 \\
\midrule
\modelterm-f32c128 & 8$\times$8$\times$128 & \cmark & 0.26 \\
\bottomrule
\end{tabular}
\caption{\textbf{Image Reconstruction Results on ImageNet 256$\times$256.} \modelfull maintains the same rFID as DC-AE under the same setting (f32c128). In addition, \modelfull has the latent space structure, while DC-AE does not. }
\vspace{-5pt}
\label{tab:rfid_results}
\end{table}

\subsection{Augmented Diffusion Training}
\label{subsec:augmented_diffusion_training}

Given the structured latent space, our next step is to utilize this structure to accelerate diffusion models' learning efficiency on object structure for better convergence. We achieve this with \diffusiontech, as demonstrated in Figure~\ref{fig:augmented_diffusion_training} (a). The core idea is explicitly augmenting the diffusion training with additional objectives on object latent channels. 

For example, let's consider a latent diffusion model \(\epsilon_\theta\) supervised by the noise prediction loss. Given the latent feature \(x_0\in \mathbb R^{\frac H f\times\frac W f\times c}\) and the corresponding noisy latent feature \(x_t=\alpha_t x_0+\beta_t \epsilon\), the denoising loss is defined as \(\|\epsilon-\epsilon_\theta (x_t, t)\|^2\).

At each diffusion training step, we randomly sample a latent channel number $(c'\le c, c' \in [c_1, c_2, \cdots, c])$, and leverage the corresponding mask \(\mathrm{mask}_{c,c'}\) defined in Eq. \ref{eq:mask} to modify the diffusion training loss
\begin{equation}
    \|\epsilon\cdot \mathrm{mask}_{c,c'}-\epsilon_\theta (x_t\cdot \mathrm{mask}_{c,c'}, t) \cdot \mathrm{mask}_{c,c'}\|^2.
\end{equation}

Figure~\ref{fig:augmented_diffusion_training} (b) illustrates the training curves of UViT-H \cite{bao2023all} with and without \diffusiontech on ImageNet 256$\times$256, demonstrating 6$\times$ faster convergence and better final image generation quality. 

\section{Experiments}
\label{sec:exp}

\subsection{Setups}

We use Pytorch \cite{ansel2024pytorch} to implement and train our models on NVIDIA H100 GPUs. We use FSDP and bf16 training to reduce the training time and training memory cost. Our largest model (USiT-3B) takes around 5 days to complete training on 16 NVIDIA H100 GPUs. 

For autoencoder training, we follow the same training strategy proposed in DC-AE \cite{chen2024deep}, including a low-resolution full training phase, a high-resolution latent adaptation phase, and a local refinement phase. We add the training technique discussed in Section~\ref{subsec:structured_latent_space} to add the channel-wise structure to the latent space. Specifically, our autoencoders support reconstructing with any latent channel number in $[16, 20, 24, ..., c]$, where $c$ is the largest latent channel number (e.g., 128).

We follow the same training settings as the official implementations for diffusion experiments, except that we increase the training batch size from 256 to 1024 for all DiT and SiT models to increase GPU utilization. We consider 4 settings in our experiments, including DiT \cite{peebles2023scalable}, UViT \cite{bao2023all}, SiT \cite{ma2024sit}, and USiT \cite{chen2024deep}. In addition to existing diffusion models, we build a larger version of USiT to conduct diffusion model scaling experiments, with a depth of 56, a hidden dimension of 2048, and 32 heads. We refer to this model as USiT-3B. By default, we do not use classifier-free guidance \cite{ho2022classifier} except stated explicitly. Evaluation metrics include FID \cite{martin2017gans}, Inception Score \cite{salimans2016improved}, Precision \cite{kynkaanniemi2019improved}, and Recall \cite{kynkaanniemi2019improved}. We profile the training throughputs of diffusion models on the NVIDIA H100 GPU with PyTorch.

\begin{table}[t]
\small\centering\setlength{\tabcolsep}{2pt}
\begin{tabular}{l | g | g | g g }
\toprule
\multicolumn{5}{l}{\textbf{ImageNet 256$\times$256}} \\
\midrule
\rowcolor{white} Diffusion & & Patch & & Inception \\
\rowcolor{white} Model & \multirow{-2}{*}{Autoencoder} & Size & \multirow{-2}{*}{gFID $\downarrow$} & Score $\uparrow$ \\
\midrule
\rowcolor{white} & DC-AE-f32c128 \cite{chen2024deep}                     & 1 & 26.44 &  53.41 \\
\multirow{-2}{*}{DiT-XL \cite{peebles2023scalable}} & \modelterm-f32c128 & 1 & \textbf{17.31} &  \textbf{80.38} \\
\midrule
\rowcolor{white} & DC-AE-f32c128 \cite{chen2024deep}                     & 1 & 17.38 &  78.42 \\
\multirow{-2}{*}{UViT-H \cite{bao2023all}} & \modelterm-f32c128          & 1 & \textbf{10.82} & \textbf{109.23} \\
\midrule
\rowcolor{white} & DC-AE-f32c128 \cite{chen2024deep}                     & 1 & 20.81 &  68.69 \\
\multirow{-2}{*}{SiT-XL \cite{ma2024sit}} & \modelterm-f32c128           & 1 & \textbf{14.91} &  \textbf{93.78} \\
\midrule
\rowcolor{white} & DC-AE-f32c128 \cite{chen2024deep}                     & 1 &  8.45 & 121.09 \\
\multirow{-2}{*}{USiT-H \cite{chen2024deep}} & \modelterm-f32c128        & 1 &  \textbf{6.10} & \textbf{146.03} \\
\bottomrule
\end{tabular}
\vspace{-5pt}
\caption{\textbf{Comparison with DC-AE on ImageNet 256$\times$256 Image Generation.} Faster convergence leads to better final results. \modelterm-f32c128 outperforms DC-AE-f32c128 under all settings. }
\vspace{-5pt}
\label{tab:better_convergence_imagenet_256}
\end{table}

\subsection{Main Results}

\paragraph{Accelerating Diffusion Model Convergence.} We compare \modelfull with DC-AE \cite{chen2024deep} under the same settings (f32c128 for ImageNet 256$\times$256 and f64c128 for ImageNet 512$\times$512) to evaluate \modelfull's effectiveness. 

Table~\ref{tab:better_convergence_imagenet_256} summarizes the class-conditional image generation results on ImageNet 256$\times$256 \cite{deng2009imagenet}. \modelterm-f32c128 not only converges faster than DC-AE-f32c128 but also results in better image generation quality, thanks to the improved convergence. It consistently delivers better gFID and Inception Scores than DC-AE-f32c128 under all settings. For example, on UViT-H \cite{bao2023all}, it improves gFID from 17.38 to 10.82 and Inception Score from 78.42 to 109.23. 

\begin{table}[t]
\small\centering\setlength{\tabcolsep}{2pt}
\begin{tabular}{l | g | g | g g }
\toprule
\multicolumn{5}{l}{\textbf{ImageNet 512$\times$512}} \\
\midrule
\rowcolor{white} Diffusion & & Patch & & Inception \\
\rowcolor{white} Model & \multirow{-2}{*}{Autoencoder} & Size & \multirow{-2}{*}{gFID $\downarrow$} & Score $\uparrow$ \\

\midrule
\rowcolor{white} & DC-AE-f64c128 \cite{chen2024deep}                     & 1 & 21.31 &  71.17 \\
\multirow{-2}{*}{DiT-XL \cite{peebles2023scalable}} & \modelterm-f64c128 & 1 & \textbf{15.16} & \textbf{95.69} \\
\midrule
\rowcolor{white} & DC-AE-f64c128 \cite{chen2024deep}                     & 1 & 17.34 &  88.47 \\
\multirow{-2}{*}{UViT-H \cite{bao2023all}} & \modelterm-f64c128          & 1 & \textbf{10.32} & \textbf{121.08} \\
\midrule
\rowcolor{white} & DC-AE-f64c128 \cite{chen2024deep}                     & 1 & 16.83 &  89.17 \\
\multirow{-2}{*}{SiT-XL \cite{ma2024sit}} & \modelterm-f64c128           & 1 & \textbf{12.44} & \textbf{114.08} \\
\midrule
\rowcolor{white} & DC-AE-f64c128 \cite{chen2024deep}                     & 1 &  6.88 & 144.24 \\
\multirow{-2}{*}{USiT-H \cite{chen2024deep}} & \modelterm-f64c128        & 1 & \textbf{5.35} & \textbf{163.03} \\
\bottomrule
\end{tabular}
\vspace{-5pt}
\caption{\textbf{Comparison with DC-AE on ImageNet 512$\times$512 Image Generation.} \modelterm-f64c128 outperforms DC-AE-f64c128 under all settings.}
\label{tab:better_convergence_imagenet_512}
\end{table}

\begin{figure}[t]
    \centering
    \includegraphics[width=0.75\linewidth]{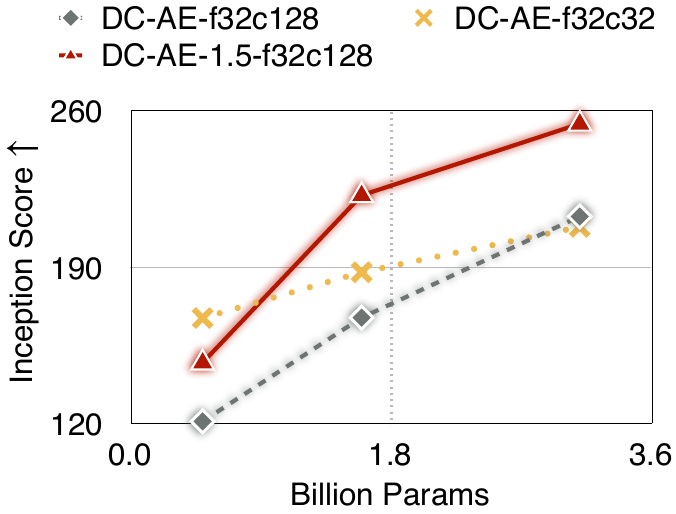}
    \vspace{-5pt}
    \caption{\textbf{Diffusion Model Scaling Results on ImageNet 256$\times$256 with USiT.} \modelterm-f32c128 delivers a better scaling curve than DC-AE-f32c32 and DC-AE-f32c128.}
    \label{fig:better_scaling_curve}
    \vspace{-10pt}
\end{figure}

\begin{table*}[t]
\small\centering\setlength{\tabcolsep}{2pt}
\begin{tabular}{l | g | g | g | g | g g | g | g | g }
\toprule
\multicolumn{9}{l}{\textbf{ImageNet 512$\times$512}} \\
\midrule
\rowcolor{white} Image & & Patch & & Training Throughput & \multicolumn{2}{c|}{gFID $\downarrow$} & Inception & & \\
\rowcolor{white} Generative Model & \multirow{-2}{*}{Autoencoder} & Size & \multirow{-2}{*}{Params (B)} & (image/s) $\uparrow$ & w/o CFG & w/ CFG & Score $\uparrow$ &  \multirow{-2}{*}{Precision $\uparrow$} & \multirow{-2}{*}{Recall $\uparrow$} \\
\midrule
\rowcolor{white} UViT-H \cite{bao2023all} & Flux-VAE-f8c16 \cite{flux2024}              & 2 & 0.50 & 55 & 30.91 & 12.63 &  56.72 & 0.64 & 0.59 \\
\rowcolor{white} DiT-XL \cite{peebles2023scalable} & Flux-VAE-f8c16 \cite{flux2024}     & 2 & 0.68 &  54 & 27.35 &  8.72 &  53.09 & 0.68 & 0.61 \\
\midrule
\rowcolor{white} DiT-XL \cite{peebles2023scalable} & SD-VAE-f8c4 \cite{rombach2022high} & 2 & 0.67 & 54 & 12.03 &  3.04 & 105.25 & 0.75 & 0.64 \\
\rowcolor{white} UViT-H \cite{bao2023all} & SD-VAE-f8c4 \cite{rombach2022high}          & 2 & 0.50 & 55 & 11.04 &  3.55 & 125.08 & 0.75 & 0.61 \\
\rowcolor{white} SiT-XL \cite{ma2024sit} & SD-VAE-f8c4 \cite{rombach2022high}           & 2 & 0.67 & 54 &     - &  2.62 &      - &    - &    - \\
\midrule
\rowcolor{white} MAGVIT-v2 \cite{yu2023language} & - & -                                & - &   - &  3.07 &  1.91 &  213.1 &    -  &    - \\
\rowcolor{white} MAR-L \cite{li2024autoregressive} & - & -                              & - &   - &  2.74 &  1.73 &  205.2  &    -  &  - \\
\rowcolor{white} EDM2-XXL \cite{karras2024analyzing} & - & -                            & - &   - &  1.91 &  1.81 &      - &         - &       -  \\
\midrule
\rowcolor{white} SiT-XL \cite{ma2024sit} & DC-AE-f32c32 \cite{chen2024deep}             & 1 & 0.67 & 241 &  7.47 &  2.41 & 131.37 & 0.77 & 0.65 \\
\rowcolor{white} USiT-H \cite{chen2024deep} & DC-AE-f32c32 \cite{chen2024deep}          & 1 & 0.50 & 247 &  3.80 &  1.89 & 174.58 & 0.78 & 0.64 \\
\rowcolor{white} USiT-2B \cite{chen2024deep} & DC-AE-f32c32 \cite{chen2024deep}         & 1 & 1.58 & \,\,112\,\,\tikzmark{usit_2b_f32p1:a1} &  \,\,2.90\,\,\tikzmark{usit_2b_f32p1:a2} &  1.72 & 187.68 & 0.79 & 0.63 \\
\midrule
\rowcolor{white} USiT-H \cite{chen2024deep} & DC-AE-f64c128 \cite{chen2024deep}         & 1 & 0.50 & 984 &  6.88 &  2.90 & 144.24 & 0.77 & 0.62 \\
\rowcolor{white} USiT-2B \cite{chen2024deep} & DC-AE-f64c128 \cite{chen2024deep}        & 1 & 1.58 & 450 &  3.20 &  2.33 & 195.94 & 0.80 & 0.61 \\
\midrule
USiT-2B \cite{chen2024deep} & \modelterm-f64c128                                        & 1 & 1.58 & \,\,450\,\,\tikzmark{usit_2b_f64p1:a1} & 2.18\tikzmark{usit_2b_f64p1:a2} & 1.84 & 237.11 & 0.80 & 0.62 \\
USiT-3B \cite{chen2024deep} & \modelterm-f64c128                                        & 1 & 3.11 & 214 & 1.70 & \,\,\,1.63$^\dag$ & 262.04 & 0.80 & 0.61 \\
\bottomrule
\end{tabular}

\begin{tikzpicture}[overlay, remember picture, shorten >=.5pt, shorten <=.5pt, transform canvas={yshift=.25\baselineskip}]

\draw [->, red] ({pic cs:usit_2b_f32p1:a1}) [bend left] to node [below right] (usit_2b_f32p1:t1) {\hspace{-2pt}\scriptsize \textbf{4$\times$ faster}} ({pic cs:usit_2b_f64p1:a1});

\draw [->, red] ({pic cs:usit_2b_f32p1:a2}) [bend left] to node [below right] (usit_2b_f32p1:t2) {\hspace{-6pt}\scriptsize \textbf{-0.72}} ({pic cs:usit_2b_f64p1:a2});

\end{tikzpicture}

\vspace{-5pt}
\caption{\textbf{Comparison with State-of-the-Art Image Generative Models on ImageNet 512$\times$512 Class-Conditional Image Generation.} \modelterm-f64c128 outperforms DC-AE-f32c32, delivering 4$\times$ higher training throughput and better image generation quality. $^\dag$ denotes the result obtained by using Guidance Interval \cite{kynkaanniemi2024applying}. }
\vspace{-10pt}
\label{tab:sota_imagenet}
\end{table*}

Table~\ref{tab:better_convergence_imagenet_512} reports the class-conditional image generation results on ImageNet 512$\times$512. The results are consistent with our findings on ImageNet 256$\times$256. \modelterm-f64c128 outperforms DC-AE-f64c128 for all cases. It again justifies \modelfull's effectiveness in accelerating diffusion model convergence and improving image generation quality for autoencoder settings where we target using a large latent channel number (e.g., c128).

\paragraph{Improving Diffusion Model Scaling Curve.} With \modelfull addressing the slow convergence issue, an exciting thing is to boost the quality upper bound of latent diffusion models by using an autoencoder with a higher latent channel number to achieve a better image reconstruction quality. To demonstrate this, we conduct diffusion model scaling experiments with USiT \cite{chen2024deep} on ImageNet 256$\times$256 and summarize the results in Figure~\ref{fig:better_scaling_curve}. 

Using DC-AE-f32c32 \cite{chen2024deep}, we can see that the inception score improvements from scaling up the diffusion models saturate due to the autoencoder's limited reconstruction quality. In contrast, scaling up diffusion models with DC-AE-f32c128 brings more significant improvements. However, due to the slow convergence, its generation quality remains inferior to DC-AE-f32c32 until the model is scaled to USiT-3B. Last, using \modelterm-f32c128, we improve the scaling curve by accelerating convergence. \modelterm-f32c128 can achieve superior Inception Score than DC-AE-f32c32 on USiT-2B and USiT-3B. 

\paragraph{Comparison with State-of-the-Art Image Generative Models on ImageNet 512$\times$512.} In addition to improving the quality upper bound of latent diffusion models, another exciting direction is to push their efficiency frontier by increasing the spatial compression ratio, such as from f32c32 to f64c128. \modelfull contributes to this direction by improving f64c128's image generation results with faster convergence. To demonstrate this, we train the USiT model (USiT-2B) with \modelterm-f64c128 on ImageNet 512$\times$512 and compare our results with state-of-the-art diffusion models and autoregressive image generative models. 

\begin{figure*}[t]
    \centering
    \includegraphics[width=\linewidth]{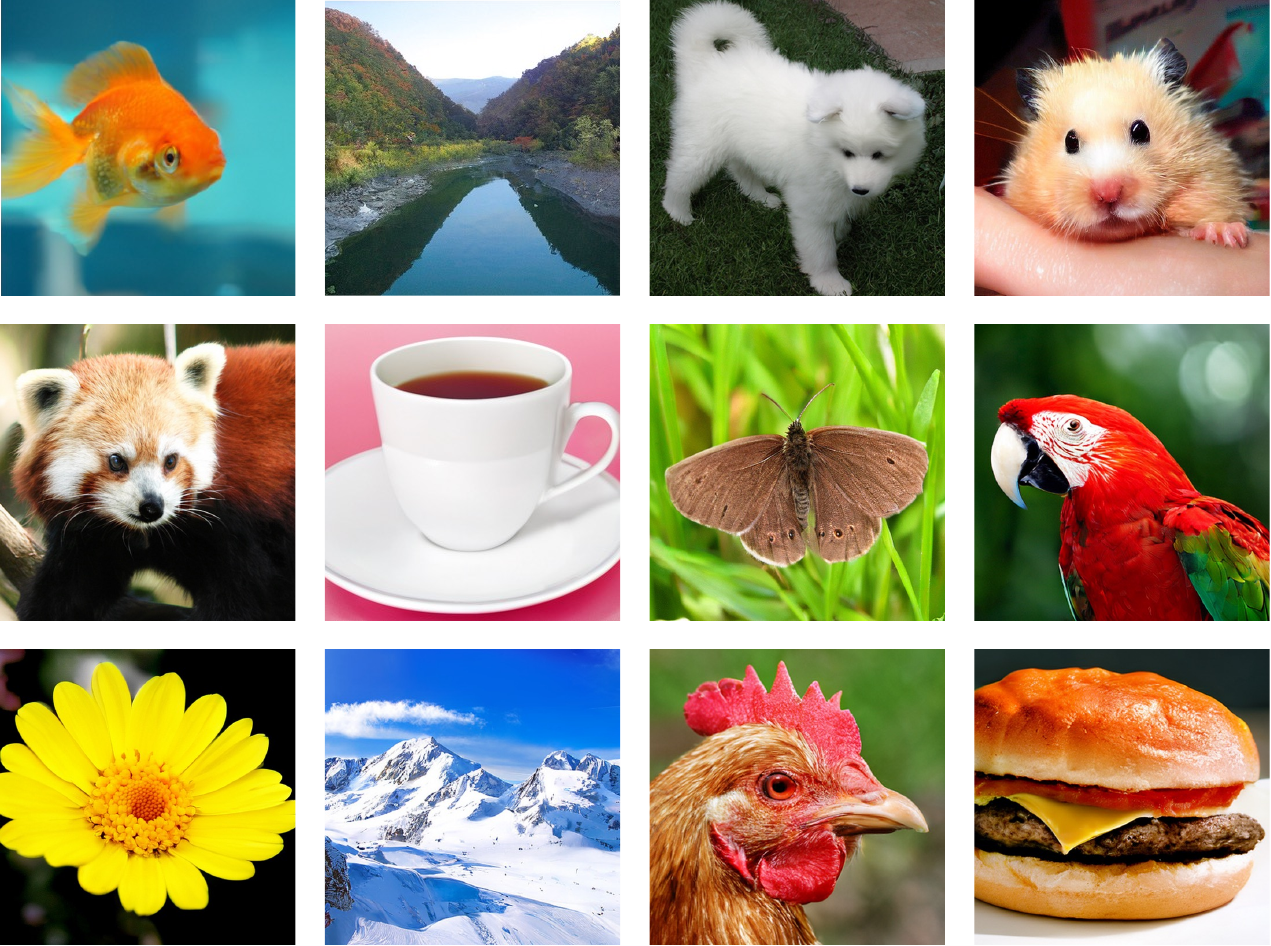}
    \vspace{-15pt}
    \caption{\textbf{Selected Image Samples Generated by Diffusion Models using \modelfull.}}
    \vspace{-5pt}
    \label{fig:generation_samples}
\end{figure*}

We summarize the results in Table~\ref{tab:sota_imagenet}. With \modelterm-f64c128, USiT-2B achieves a competitive image generation quality with exceptional efficiency. In particular, it achieves a gFID of 2.18, an inception score of 237.11 without using classifier-free guidance, significantly outperforming DC-AE-f32c32+USiT-2B. More importantly, it delivers 4$\times$ higher training throughput than DC-AE-f32c32+USiT-2B. 
Figure~\ref{fig:generation_samples} demonstrates our image generation examples, showing competitive visual quality compared with other generative models trained on ImageNet. 

\begin{table}[t]
\small\centering\setlength{\tabcolsep}{2pt}
\resizebox{1\linewidth}{!}{
\begin{tabular}{l | g g | g g }
\toprule
\multicolumn{5}{l}{\textbf{ImageNet 256$\times$256}} \\
\midrule
\rowcolor{white} Diffusion & Structured & Augmented & & Inception \\
\rowcolor{white} Model & Latent Space & Diffusion Training & \multirow{-2}{*}{gFID $\downarrow$} & Score $\uparrow$ \\
\midrule
\rowcolor{white} & &                                                  & 26.44 &  53.41 \\
\rowcolor{white} & \cmark &                                           & 26.75 &  51.07 \\
\rowcolor{white} & & \cmark                                           & 36.83 &  42.22 \\
\multirow{-4}{*}{DiT-XL \cite{peebles2023scalable}} & \cmark & \cmark & \textbf{17.31} &  \textbf{80.38} \\
\midrule
\rowcolor{white} & &                                                  &  8.45 & 121.09 \\
\rowcolor{white} & \cmark &                                           &  6.72 & 132.02 \\
\rowcolor{white} & & \cmark                                           &  8.99 & 125.89 \\
\multirow{-4}{*}{USiT-H \cite{chen2024deep}} & \cmark & \cmark        &  \textbf{6.10} & \textbf{146.03} \\
\bottomrule
\end{tabular}
}
\vspace{-5pt}
\caption{\textbf{Component-wise Ablation Study Experiments.} Removing either \aetech or \diffusiontech causes a significant image generation quality drop, showing it is critical to use these two techniques together. }
\vspace{-10pt}
\label{tab:ablation_components}
\end{table}

\subsection{Ablation Study}

\paragraph{Component-Wise Analysis of \modelfull.} We conduct component-wise ablation to investigate the impact of \modelfull's two key designs, i.e., \aetech (Section~\ref{subsec:structured_latent_space}) and \diffusiontech (Section~\ref{subsec:augmented_diffusion_training}). If \aetech is enabled, we employ \modelterm-f32c128; otherwise, we use DC-AE-f32c128.

\begin{table}[t]
\small\centering\setlength{\tabcolsep}{3pt}
\begin{tabular}{l | c c | c c }
\toprule
\multicolumn{4}{l}{\textbf{ImageNet 256$\times$256}} \\
\midrule
Diffusion & & & & Inception \\
Model & \multirow{-2}{*}{Setting} & \multirow{-2}{*}{Autoencoder} & \multirow{-2}{*}{gFID $\downarrow$} & Score $\uparrow$ \\
\midrule
& & DC-AE \cite{chen2024deep}                                                               & 10.18 & 107.49 \\
\multirow{-2}{*}{DiT-XL \cite{peebles2023scalable}} & \multirow{-2}{*}{f32c32} & \modelterm & 10.50 & 107.99  \\
\midrule
& & DC-AE \cite{chen2024deep}                                                               &  3.89 & 167.40 \\
\multirow{-2}{*}{USiT-H \cite{chen2024deep}} & \multirow{-2}{*}{f32c32} & \modelterm        &  4.15 & 166.94 \\
\bottomrule
\end{tabular}
\vspace{-5pt}
\caption{\textbf{Comparison with DC-AE under f32c32.}}
\vspace{-5pt}
\label{tab:ablation_comparison_f32c32}
\end{table}

We report the results in Table~\ref{tab:ablation_components}. We can see that both \aetech and \diffusiontech are essential for improving the image generation quality. For example, on DiT-XL \cite{peebles2023scalable}, adding \aetech or \diffusiontech alone even hurts the results. In contrast, combining them leads to clear improvements, decreasing the gFID from 26.44 to 17.31 and increasing the inception score from 53.41 to 80.38. Thus, we recommend using these two techniques together. 

\paragraph{Comparison with DC-AE under f32c32.} While \modelfull is mainly for autoencoder settings with large latent channel numbers (e.g., c128), we can still apply it to settings with small latent channel numbers (e.g., c32). Table~\ref{tab:ablation_comparison_f32c32} summarizes the comparison between \modelfull and DC-AE under f32c32. We find DC-AE-f32c32 performs slightly better than \modelterm-f32c32. We conjecture that f32c32 does not have the latent space sparsity issue. Thus, adding \aetech and \diffusiontech is not necessary. It is consistent with our analysis in Section~\ref{subsec:analysis_and_motivation} and Figure~\ref{fig:analysis_and_motivation}. Based on this finding, we recommend using \modelfull when targeting a large latent channel number (e.g., c128) and conventional autoencoders such as DC-AE when targeting a small latent channel number (e.g., c32). 

\section{Conclusion}
\label{sec:conclusion}

This paper presents a new approach for addressing the slow convergence issue of latent diffusion models when using a large latent channel number. Our approach consists of two key innovations, including \aetech for imposing a desired channel-wise structure on the latent space and \diffusiontech to utilize the latent space structure to accelerate object structure learning. These techniques collectively enable faster diffusion model convergence and lead to better image generation results, paving the way for employing autoencoders with higher quality upper bound or higher spatial compression ratios.

{
    \small
    \bibliographystyle{ieeenat_fullname}
    \bibliography{main}
}

\newpage
\onecolumn
{
\centering
\Large
\textbf{\thetitle}\\
\vspace{0.5em}Supplementary Material \\
\vspace{1.0em}
}

\setcounter{page}{1}
\setcounter{section}{0}
\renewcommand*{\thesection}{\Alph{section}}

\section{Improving Diffusion Model Scaling Curve}

\begin{figure}[htbp]
    \centering
    \includegraphics[width=0.375\linewidth]{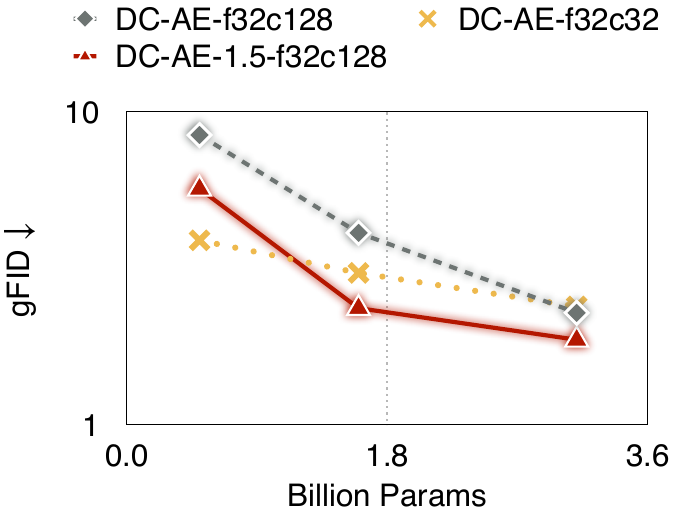}
    \caption{\textbf{Diffusion Model Scaling Results on ImageNet 256$\times$256 with USiT.} \modelterm-f32c128 delivers a better scaling curve than DC-AE-f32c32 and DC-AE-f32c128.}
    \label{fig:better_scaling_curve_fid}
\end{figure}

Besides Figure~\ref{fig:better_scaling_curve}, we show the FID metric for the diffusion model scaling experiments in Figure~\ref{fig:better_scaling_curve_fid}. The observation is similar, the generation quality of diffusion models using DC-AE-f32c128 remains inferior to DC-AE-f32c32 until the model is scaled to USiT-3B, while \modelterm-f32c128 can achieve superior Inception Score than DC-AE-f32c32 on USiT-2B and USiT-3B. 

\section{Evaluation Details}

We follow the common practice to evaluate our autoencoders and latent diffusion models on the ImageNet \cite{deng2009imagenet} dataset.

For image reconstruction experiments, we compute the metrics using the 50,000 validation images and their reconstruction results. 

For image generation experiments, we generate 50,000 images and compute the metrics with the training split.

The evaluation metrics include FID \cite{martin2017gans}, Inception Score \cite{salimans2016improved}, Precision \cite{kynkaanniemi2019improved}, and Recall \cite{kynkaanniemi2019improved}. The specific implementation of these metrics are provided in our codebase.

\section{Complete Latent Space Visualization}

\begin{figure*}[h]
    \centering
    \includegraphics[width=\linewidth]{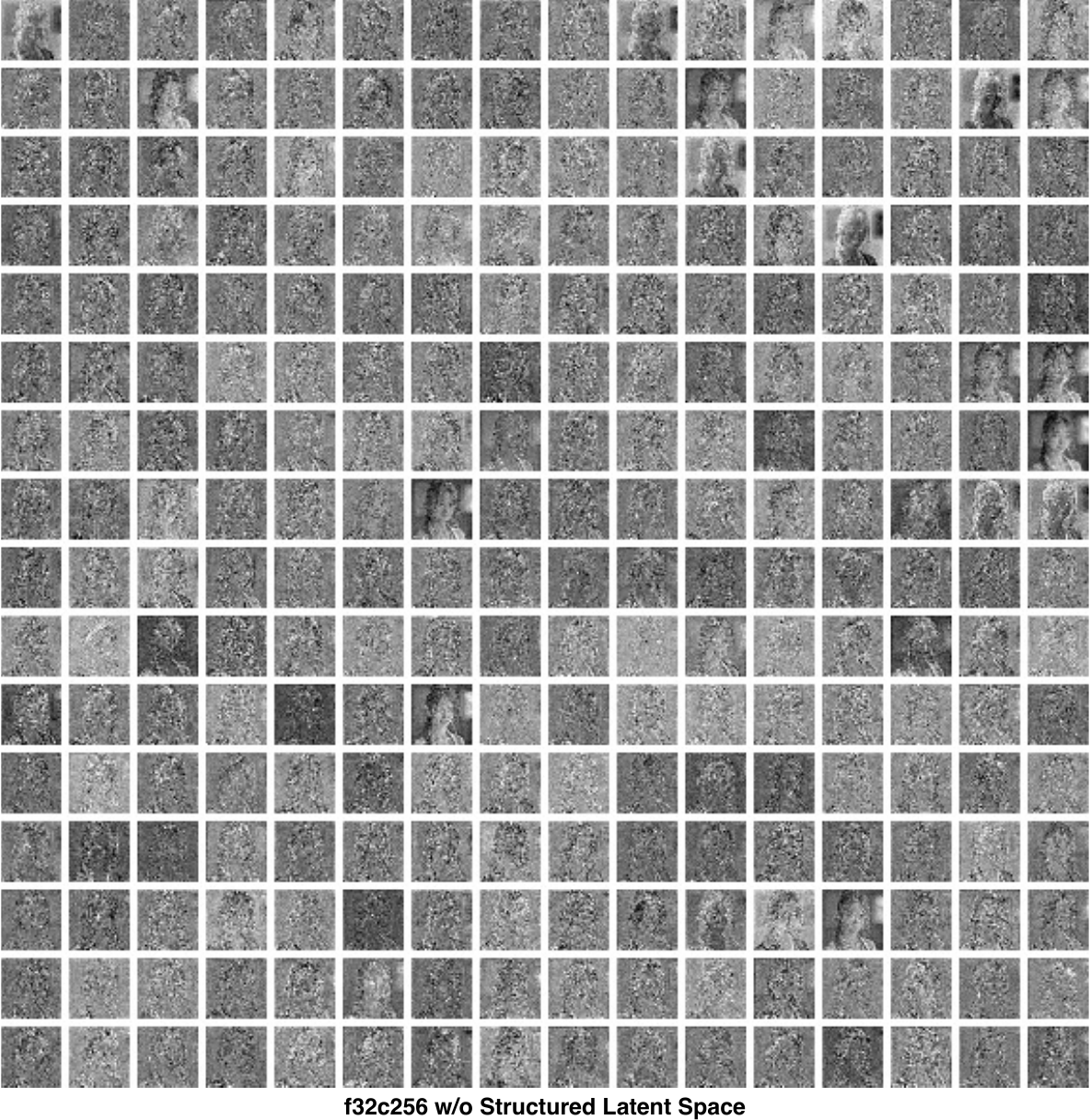}
    \caption{\textbf{Complete Latent Space Visualization of DC-AE-f32c256.}}
    \label{fig:latent_space_visualization_1}
\end{figure*}

\begin{figure*}[t]
    \centering
    \begin{subfigure}[t]{1.0\linewidth}
        \centering
        \includegraphics[width=1.0\linewidth]{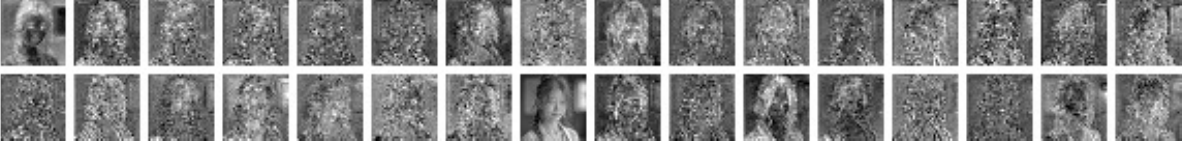}
        \caption{DC-AE-f32c32}
    \end{subfigure}
    \begin{subfigure}[t]{1.0\linewidth}
        \centering
        \includegraphics[width=1.0\linewidth]{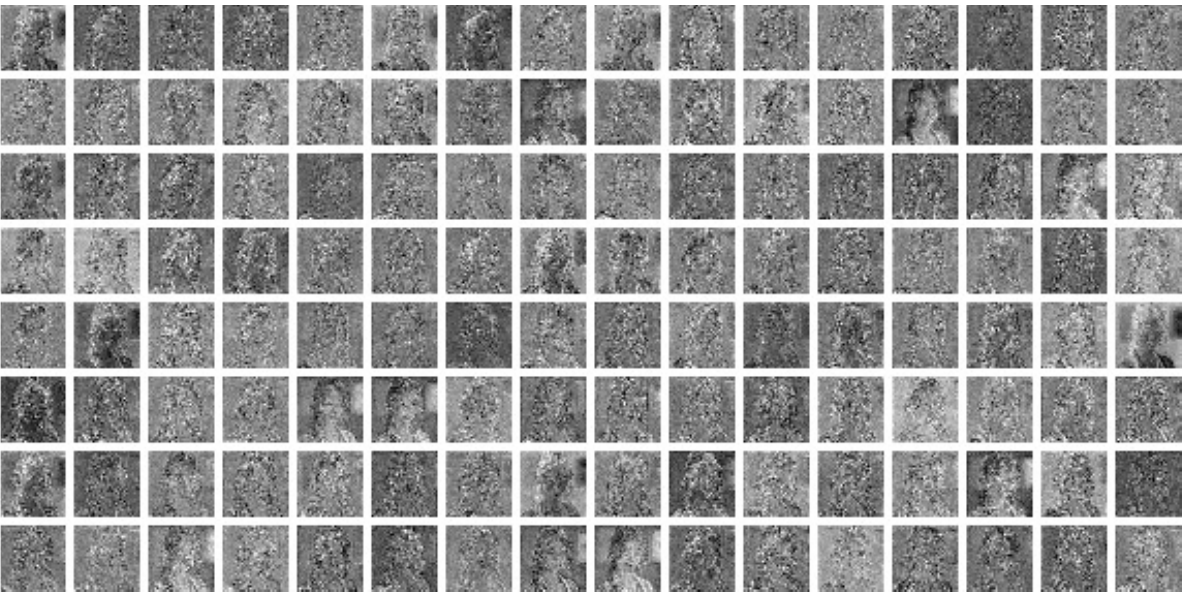}
        \caption{DC-AE-f32c128}
    \end{subfigure}
    \begin{subfigure}[t]{1.0\linewidth}
        \centering
        \includegraphics[width=1.0\linewidth]{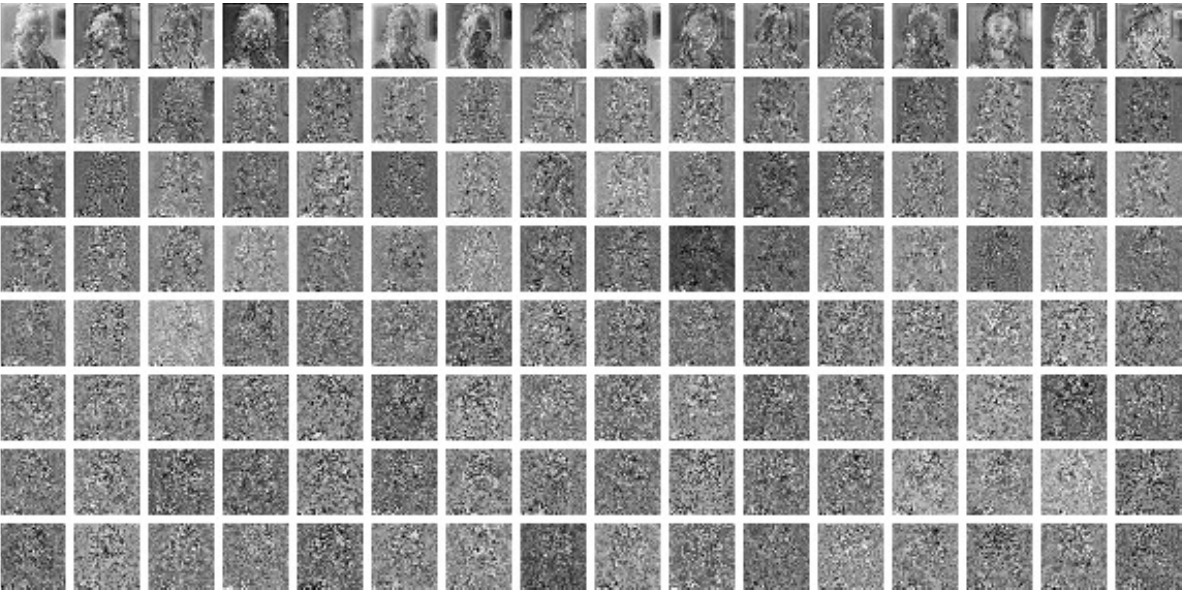}
        \caption{\modelterm-f32c128 with structured latent space}
    \end{subfigure}
    \caption{\textbf{Complete Latent Space Visualization of DC-AE-f32c32, DC-AE-f32c128, and \modelterm-f32c128.}}
    \label{fig:latent_space_visualization_2}
\end{figure*}


\end{document}